\documentclass[11pt,a4paper]{article}
\usepackage[hyperref]{emnlp2020}
\usepackage{times}
\usepackage{diagbox}
\usepackage{latexsym}

\usepackage{times}  
\usepackage{helvet}  
\usepackage{courier}  
\usepackage{url}  
\usepackage{graphicx}  
\usepackage{courier}
\usepackage{amsmath}
\usepackage{graphicx}
\usepackage{multirow}
\usepackage{mathbbol}
\usepackage{caption}
\usepackage{CJKutf8}
\usepackage{verbatim}
\usepackage{float}
\usepackage{booktabs}
\usepackage{array}
\usepackage{times}
\usepackage{csquotes}
\usepackage{latexsym}
\usepackage{tablefootnote}
\usepackage[normalem]{ulem}
\usepackage{todonotes}
\usepackage{subcaption}
\useunder{\uline}{\ul}{}

\usepackage{microtype}

\aclfinalcopy 
\def\aclpaperid{****} 

\title{Identifying Distributional Perspective Differences from Colingual Groups}

\author{Yufei Tian$^1$
, Tuhin Chakrabarty$^{2}$, Fred Morstatter$^{3}$ and Nanyun Peng$^{3,4}$\\
  $^1$ Department of Computer Science, University of Southern California\\
  $^2$ Department of Computer Science, Columbia University\\
  $^3$ Information Sciences Institute, University of Southern California\\
  $^4$ Department of Computer Science,  University of California Los Angeles\\
  \texttt{yufeit@usc.edu, } \texttt{tuhin.chakr@cs.columbia.edu,} \\ \texttt{fredsmors@isi.edu,} \texttt{violetpeng@cs.ucla.edu} }

\date{}

\begin{document}
\maketitle
\begin{abstract}

Perspective differences exist among different cultures or languages. A lack of mutual understanding among different groups about their perspectives on specific values or events may lead to uninformed decisions or biased opinions. Automatically understanding the group perspectives can provide essential background for many downstream applications of natural language processing techniques. In this paper, we study colingual groups\footnote{A group of people that share the same language (https://www.merriam-webster.com/dictionary/colingual).} and use language corpora as a proxy to identify their \textit{distributional perspectives}. We present a novel computational approach to learn shared understandings, and benchmark our method by building culturally-aware models for the English, Chinese, and Japanese languages. On a held out set of diverse topics including \textit{marriage}, \textit{corruption}, \textit{democracy}, our model achieves high correlation with human judgements regarding intra-group values and inter-group differences.



\end{abstract}

\section{Introduction}
\label{sec:intro}


\begin{table}[t!]
\centering
\small
\begin{subtable}{\columnwidth}
\begin{tabular}{|@{ }p{1.2cm}@{ }|@{ }p{6.1cm}@{ }|}
\hline
Claim  & \begin{tabular}[c]{@{}p{6.1cm}@{}}The free market does a much worse job than the government in providing essential services and the fraud and corruption part only gets worse.\end{tabular} \\ \hline
CN Persp & Human: 72\% support, Model: 79\% support\\ \hline
JP Persp & Human: 17\% support, Model: 15\% support\\ \hline
\end{tabular}
\caption{A claim about free market and government intervention from our test data, with the distributional perspectives of the Chinese (CN) and Japanese (JP) colingual groups. Human opinions and model predictions are highly correlated.}\label{tab:claim} 
\end{subtable}

\begin{subtable}{\columnwidth}
\begin{tabular}{|@{ }p{1.2cm}@{ }|@{ }p{6.1cm}@{ }|}
\hline

\multirow{2}{*}{\begin{tabular}[c]{@{}p{1.2cm}@{}}CN \\ Wiki- \\pedia \end{tabular}}  & \begin{CJK}{UTF8}{gkai} \begin{tabular}[c]{@{}p{6.0cm}@{}}中国特色的社会主义现阶段有如下特点: 以国家的手段控制国内的要害经济部门和大量的企业，通过``国有资产''的概念以股份或者非股份形式保护国民经济的相当重要的部分。\end{tabular}  \end{CJK} \\ \cline{2-2} 
  & \begin{tabular}[c]{@{}p{6.2cm}@{}}
  The current stage of socialism with Chinese characteristics has the following characteristics: the government \textit{\color{red}control the vital economic sectors} and a large number of enterprises in the country \textit{\color{red}by state means}, and protect a very important part of the national economy in the form of shares or non-shares through the concept of \textit{\color{red}``state-owned assets''}.
   {[}Translated{]}\end{tabular} \\ \hline
\multirow{2}{*}{\begin{tabular}[c]{@{}p{1.1cm}@{}} JP \\Wiki- \\pedia \end{tabular}} &\begin{CJK}{UTF8}{gkai} \begin{tabular}[c]{@{}p{6.0cm}@{}}1930年以降、社会的市場経済に対して個人の自由や市場原理を再評価し、政府による個人や市場への介入は最低限とすべきと提唱する。... 日本では1950年の電気事業再編成以来、民営の電力会社が地域ごとに1社ずつ合計10社あり。 \end{tabular}    \end{CJK}\\ \cline{2-2} 
  & \begin{tabular}[c]{@{}p{6.0cm}@{}} Since 1930, Japan reassessed the liberty and market principles of the individual for the social market economy, \textit{\color{blue}advocating that government intervention} in the individual and \textit{\color{blue}the market} \textit{\color{blue}should be minimized}. ... In Japan, since the restructuring of the electric power business in 1950, \textit{\color{blue}there are 10 private electric power companies, one in each region}.  [Translated]\end{tabular} \\ \hline
\end{tabular}
\caption{Evidence from Wikipedia pages from the colingual groups (CN and JP), that potentially are for or against the claim shown in Table~\ref{tab:claim}. These are included in our training data after variation (discussed in Section \ref{sec:bias}). The two examples in the JP corpus are selected out from different articles.}\label{tab:wiki}
\vspace{-.5em}
\end{subtable}
\caption{\label{tab:1} An example claim from our test data (\ref{tab:claim}), and possible evidences from wikipedia pages included in our colingual group training corpora (\ref{tab:wiki}).}
\vspace{-2em}
\end{table}


Sociologists have defined culture as a set of shared understandings, herein called \textit{perspectives}, adopted by the members of that culture~\cite{bar2000shared,sperber2004foundations}. Languages and cultures have radical correlations~\cite{khaslavsky1998integrating,bracewell2012language,gelman2017language}, because individuals communicate with each other by language, which carries the aspects of their cultures, experiences, beliefs, and values, thus will shape their perspectives.
Lacking of understanding for these perspective differences could lead to biased predictions. Selection bias \cite{heckman1977sample} can often lead to misinformation as it sometimes ignores facts that do not reflect the entire population intended to be analyzed. For example, to verify a controversial statement like \textit{``The free market causes fraud and corruption."}, we need to consider the perspectives from various groups (shown in Table \ref{tab:1}). Similarly, a sentiment analysis model may fail to capture the correct emotions towards a debatable claim if the claim is viewed differently across different groups, such as the dispute between India and Pakistan regarding Kashmir.



In this paper, we focus on \textit{distributional differences} on controversial topics across groups. For example, within the United States, people have split views (approximately half-half) regarding gun control and abortion, while in China, people generally against the possession of guns and pro-choice for abortion. Hence, building a culture-aware model that considers groups' \textit{distributional} perspectives will help improve comprehension and consequentially mitigate biases in decision making. 

We aim to identify colingual groups' distributional perspectives towards a given claim, and spot claims that provoke such divergence. As colingual groups are naturally identifiable by the usage of language, we can obviate group detection and associated errors in the process of group identification.\footnote{The only caveat is that such simplification ignores finer-grained cultural distinctions across subgroups speaking the same language, especially for English as a global language spoken by many nations; we leave those studies of more fine-grained groups for future work.} Wikipedia, despite its overall goal of objectivity, has been shown to embed latent cultural biases~\cite{callahan2011cultural}. Following these cues, we believe Wikipedia is an ideal source to study diverse perspectives among various colingual groups. Table~\ref{tab:claim} shows an example claim for which the Chinese and Japanese may have different opinions. Specifically, the Chinese-speaking group tends to support the claim (72\% support) while the Japanese-speaking group tends to oppose it (17\% support), which is likely due to the different economic/government environments. As shown in Table~\ref{tab:wiki}, we can find evidences from wikipedia pages that support or oppose the claim in Table~\ref{tab:claim}.

We learn a perspective model for each colingual groups using a collection of Wikipedia pages for English, Chinese and Japanese, and then use these models to identify diverging perspectives for a separate set of claims that are manually curated and are \textit{not} from Wikipedia. 

Our contributions are as follows. \textbf{1)} We propose \textbf{CLUSTER} (\textbf{C}o\textbf{L}ing\textbf{U}al Per\textbf{S}pec\textbf{T}ive Identifi\textbf{ER}), a module that learns distributional perspectives of colingual groups based on Wikipedia articles. Towards this, we develop a novel procedure to algorithmically generate negative examples (introduced in Section \ref{subsec:dataset creation}) based on Wikipedia to train our group models (Section \ref{subsec:train_proc}).
\textbf{2)} We design an evaluation framework to systematically study the effectiveness of the proposed approach by testing our models on self-labeled claims  from diverse topics including \textit{cuisine, festivals, marriage, corruption, democracy, privacy}, etc. (Section \ref{subsec:test_data}, \ref{subsec:HATD} and \ref{subsec:inference})
\textbf{3)} Comprehensive quantitative and qualitative studies in Chinese, Japanese, and English show that our model outperforms multiple well-crafted baselines and achieves strong correlation with human judgements.\footnote{We use these three languages as examples throughout the paper, but our algorithm is naturally applicable to other languages. Data and code are available at https://github.com/PlusLabNLP/CLUSTER} (Section \ref{sec:results} and \ref{sec:quanlitative_analysis})



\section{Task Definition}
\label{sec:task}
In this paper, we focus on predicting a group's distributional perspective towards a \textit{claim} and identifying claims that reflect \textit{contrasting perspectives} from different groups on a particular topic. We further focus on English, Chinese and Japanese as the targeted colingual groups. Here, we define several key concepts and the task. We also explain why our task is different from stance detection.

\paragraph{Claim.}
A claim $s_i$, is a sentence that expresses opinions toward a certain topic (E.g Row 1 , Table \ref{tab:1}) regardless of its language.  We then translate and have a set of multi-lingual claims $\mathcal{S} = (\mathcal{S}^{en}, \mathcal{S}^{cn}\,\mathcal{S}^{jp})$, where $\mathcal{S}^{en} $ (English), $\mathcal{S}^{cn}$ (Chinese), $\mathcal{S}^{jp}$ (Japanese) are translations of each other.  




\paragraph{Group Perspective Model and Score.}
Group Perspective Model is a probabilistic model that mirrors the group's distributional perspective on a claim - the model gives a score that reflects a group's likelihood of agreeing with that claim. For any claim $s$ and its translations $(s^{en},s^{cn},s^{jp}),$ a machine-generated score $P^l(s^{l}) \in [0,1]$ is assigned to estimate the probability of  $s^{l}$ ($l$ denoting language) being supported by the corresponding group. A distributional perspective score closer to 1 (fully support) and 0 (fully reject) indicates unanimity, while a score closer to 0.5 implies split within group. Similarly, a human-annotated perspective score $H^l(s^{l}) \in [0,1]$ is assigned and considered as the ground truth of the likelihood that $s^{l}$ is supported by its corresponding group.



\paragraph{Distributional Perspective Difference.}
Finally, we define (distributional) perspective difference.
Let $\mathcal{D}_{model_i}^{l_1-l_2} \in [-1,1]$ be the difference of perspective scores predicted by two models (for group $l_1$ and $l_2$) of $s$, where

\vspace*{-4mm}
\begin{equation}
    \mathcal{D}_{model}^{l_1-l_2} =  P^{l_1}(s^{l_{1}})-P^{l_2}(s^{l_{2}}),  l_1\neq l_2.
\label{eq:model}
\end{equation}
Here $l_1$ and $l_2$ each denotes a language such as `cn' and `jp'. A positive $\mathcal{D}_{model}^{cn-jp}$ indicates that the Chinese model agrees more with the claim $s$ than the Japanese model. Similarly, we denote $\mathcal{D}_{human}^{l_1-l_2} \in [-1,1]$ as the quantity of perspective difference reported by human annotators:

 \vspace*{-4mm}
 \begin{equation}
     \mathcal{D}_{human}^{l_1-l_2} =  H^{l_1}(s^{l_{1}})-H^{l_2}(s^{l_{2}}),  l_1\neq l_2.
     \label{eq:human}
 \end{equation}

In Table \ref{tab:1}, $\mathcal{D}_{model}^{cn-jp} = 0.79-0.15 = 0.64$, and $\mathcal{D}_{human}^{cn-jp} = 0.55$. A higher absolute value of $\mathcal{D}$ indicates bigger distributional differences. 


\begin{figure*}[ht!]
\centering
\includegraphics[width=0.9\textwidth]{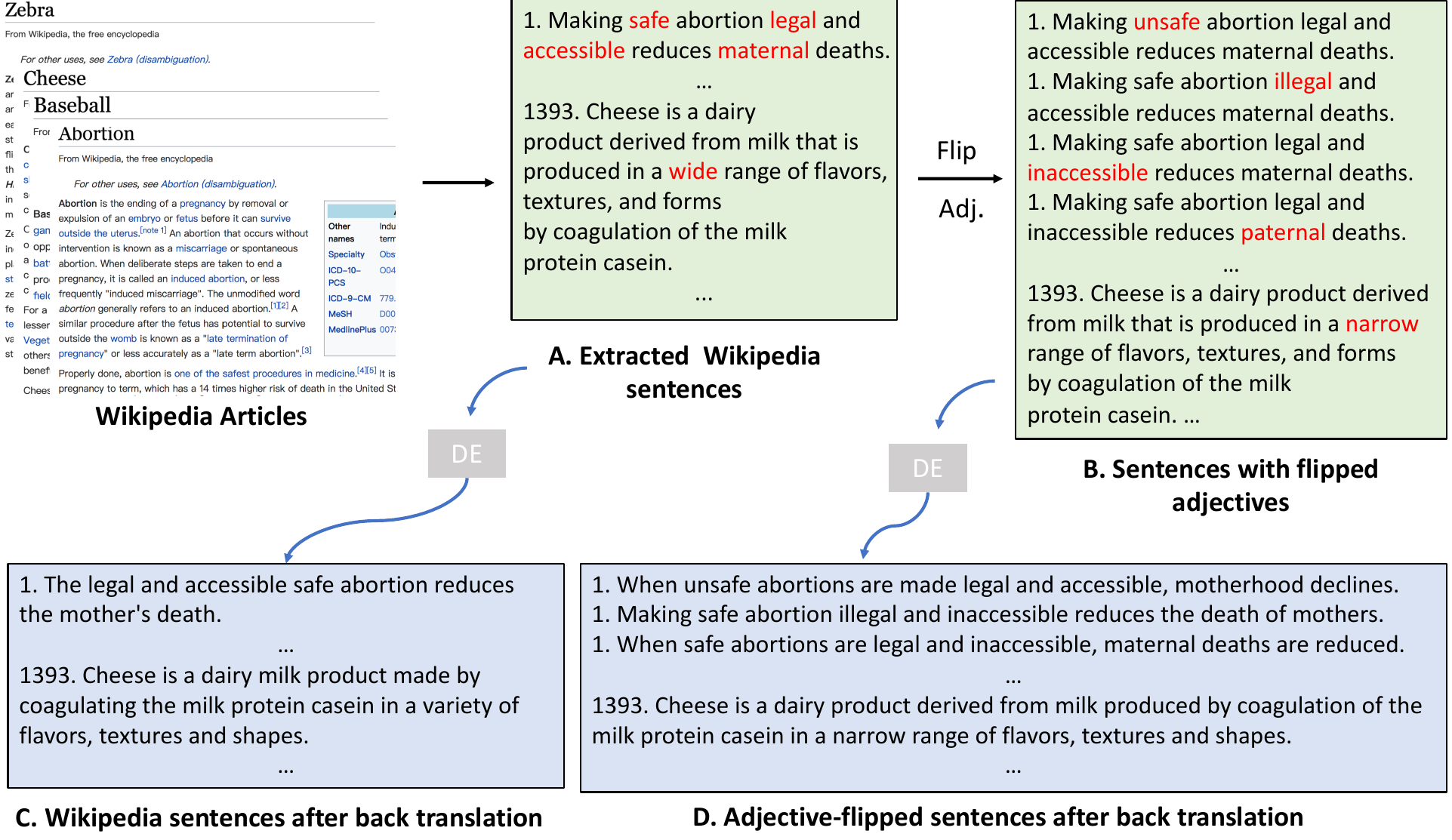}
\caption{An illustration of the creation of the English training data. We first extract sentences from the retrieved Wikipedia articles to form the positive samples, and then replace adjectives with their antonyms as negative samples. \textit{Back-translation} (discussed in \ref{sec:bias}) is then used to resolve pattern bias among negative samples. Note that we do not flip multiple adjectives simultaneously.}
\label{fig::training_data}
\vspace*{-3mm}
\end{figure*}

\paragraph{Comparison With Stance Detection.}
Stance detection aims at detecting if a piece of text (usually a sentence or a document) supports or opposes a given claim~\cite{hasan-ng-2014-taking}. Unlike stance detection, we do not have a given text associated with our claims. Instead, we learn representations of \textit{group} perspectives through training on language corpora so that we can identify if a claim is likely to be supported or opposed by \textit{a group}.
 

\section{Data Preparation}
\label{sec:data}

In this section, we describe the procedure of composing our training data from multi-lingual Wikipedia articles. We then introduce an out-of-domain test dataset retrieved from Reddit that contain opinions regarding wide range of topics and the procedure of collecting human annotations on the test set.

\subsection{Training Data}
 \label{sec:training-data}
\paragraph{Topic Selection.} 
\label{subsection:category tree}

 We leverage the category hierarchy provided by Wikipedia to retrieve a list of child topics that belong to a few parent categories, including \textit{{politics}}, 
 \textit{{foods}}, 
 \textit{{sport}}, 
 \textit{{history}},\textit{social issues}, etc. 
 The selected root categories in English, Chinese and Japanese are aligned entities obtained from Wikipedia language links, and their sub-tree structures are only partially aligned. In this way,   sub-topics obtained in the three languages have considerable overlap but are not identical. Hence we have different numbers of subtopics and training samples as seen in Table \ref{table:data}. We then retrieve all the articles under the selected subtopics separately\footnote{https://en.wikipedia.org/wiki/Special:CategoryTree, \begin{CJK}{UTF8}{gkai}{https://zh.wikipedia.org/w/title=Special:分类树, https://ja.wikipedia.org/wiki/特别:カテゴリツリー}\end{CJK}}, so that different claims that potentially reflect the cultural bias are included in our training data.

\paragraph{Training Dataset Creation.}
\label{subsec:dataset creation}

Upon observing  many examples similar to the economics pages in Table \ref{tab:1}, we form our fundamental assumption that the \textit{collection} of sentences extracted from Wikipedia in a certain language represent the corresponding \textit{distributional} perspective of that colingual group. Therefore, we label each sentence extracted from the Wikipedia articles as \textit{\textbf{positive examples}}, as illustrated in part A of Figure \ref{fig::training_data}.

Although positive examples mirror their corresponding perspective, we also need to compose  \textit{\textbf{negative samples}} --- the claims that the corresponding colingual groups will disagree with. 
An intuitive approach is to flip the semantic meaning of the positive examples. This could be achieved by replacing the adjectives in a sentence with their antonyms. As shown in Figure~\ref{fig::training_data}.A, there are four adjectives in the original text: \textit{`Making \textbf{safe} abortion \textbf{legal} and \textbf{accessible} reduces \textbf{maternal} deaths'}. We can obtain four negative examples by replacing each of the adjectives with its antonym (note that we do not flip multiple adjectives simultaneously). Each of the fabricated negative samples (in Figure~\ref{fig::training_data}.B) is ideal because it expresses conflicting viewpoints compared to the original text.

However, certain collocations such as \textit{New York} and \textit{legal systems} are also converted. Since bigrams such as \textit{Old York} and \textit{illegal systems} seldom appear in real sentences, we use a statistical n-gram model to avoid those poorly constructed negative samples.  
So far, we've obtained all data to train the perspective models. We list the number of topics, retrieved sentences, and training samples in Table~\ref{table:data}.

\begin{table}[t]
\centering
\small
\begin{tabular}{c|ccc}
\toprule
        & Topics & \begin{tabular}[c]{@{}c@{}}Positive\\ Samples\end{tabular} & \begin{tabular}[c]{@{}c@{}}Negative\\ Samples\end{tabular} \\ \midrule
English & 4,245     & 292,444                                                    & 292,444                                                    \\ 
Chinese & 1,563     & 57,904                                                     & 57,904                                                     \\ 
Japanese & 1,266  & 25,039    & 25,039                                                             \\ \bottomrule
\end{tabular}
\caption{Statistics of Our Training Dataset. We deliberately balance the number of positive and negative samples so that no priori probability will intervene with the learning step.}
\vspace{-1em}
\label{table:data}
\end{table}

\subsection{Out-of-domain Test Data}
\label{subsec:test_data}

While training and testing on the same Wikipedia data is a possible choice, a more ideal scenario is to test on different domains to see if the distributional representation learned by the model generalizes to other datasets, not merely representing the style of Wikipedia. 
Hence, selecting a good held-out set to test the performance of our models is important. 

We are motivated by the fact that people always express personal opinions on social media such as Reddit, where many opinionated claims are included. We leverage a previous work \cite{chakrabarty2019imho} which collects a distant supervision-labeled corpus of 5.5 million opinionated claims covering a wide range of topics using sentences containing the acronyms IMO (in my opinion) or IMHO (in my humble opinion) from Reddit. Table~\ref{table2:example} shows two examples from the IMO dataset that may reveal contrasting perspectives between two different colingual groups. As this dataset is only in English, to obtain scores from the Chinese and Japanese cultural models, we translate each sentence into the target language using the Youdao and Google Translate API\footnote{https://ai.youdao.com, https://translate.google.com}.

 \begin{table}[t]
 \centering
 \small
 \begin{tabular}{|l|}
 \hline
 \begin{tabular}[c]{@{}l@{}}\textbf{IMHO}, what I find strange, and this is totally, some\\ Chinese people have dogs as both pets and as dinner. \tablefootnote{This does not reflect the opinion of the authors.}\end{tabular}      \\ \hline
 \begin{tabular}[c]{@{}l@{}}\textbf{IMO}, in an utopia Communism is the best system to\\ live by.\end{tabular} \\ \hline
 \end{tabular}
 \caption{Sentence from the IMO dataset expressing opinions about which differ between cultures.}
 \label{table2:example}
 \vspace{-4mm}
 \end{table}

\paragraph{Test Data Selection.}
 We first automatically extract claims that contain certain topical keywords, such as \textit{free market} and \textit{democracy}, and then remove the candidates which are out-of-context. Then we ask the English and Chinese volunteers to jointly select high-quality statements. 
Finally, for human annotation, we select out 128 high-quality claims from over 2,000 candidates in the IMO/IMHO dataset. The topics include personal life, social and political views, etc.


\subsection{Human Annotations for Test Data}\label{subsec:HATD}
For each test sample, we collect 20 annotations from annotators living in the United States using the Amazon Mechanical Turk platform (MTurk). We then collect another 20 annotations from Chinese/Japanese netizens using the SurveyHero/Crowdworks\footnote{https://www.surveyhero.com, https://crowdworks.jp} platform because MTurk is less used by the local people. The annotations are binarized, with 1 indicating agreement and 0 indicating disagreement. The average scores are viewed as the distributional scores. 

For instance, for a given claim $s^{en_i}$, if 13 out of 20 English annotators give scores of 1, and the other 7 give scores of 0, then the human-annotated score $H^{en}(s^{en_i})$ equals $ 13/20 = 0.65$. In this way, we ensure that human annotation is of the same scale and meaning as the model prediction, and thus prove the validity of using the correlation between model predictions and human annotations as a measurement of effectiveness.

\section{Methodology}
\label{sec:method}
In this section, we present the procedure of training our CLUSTER model. We explain how to learn group perspective models for English, Chinese, and Japanese colingual groups. We then raise the issue of pattern bias in negative samples and provide our corresponding solution. Lastly, we introduce the inference process.

\subsection{Training Process}
\label{subsec:train_proc}
In the training stage, we leverage the pretrained multilingual BERT \citep{devlin2018bert} and fine-tune it for the perspective-specific classification task on the labeled data that is obtained in \ref{subsec:dataset creation}.


To enable the whole system to capture as much cultural discrepancy as possible, we separately fine-tune a BERT model for each language corpora despite the multilingualism of BERT. In other words, the learning steps of English, Chinese and Japanese systems have  exactly the same structure but are completely isolated from each other in terms of training data and model parameters.

\subsection{Pattern Bias in Negative Samples and Targeted Improvements}
\label{sec:bias}

While flipping adjectives to create negative samples appears as an obvious approach, it ends up introducing certain style biases. Since the placeholders for adjectives are the only difference between positive and negative samples in training data, most classifiers would be able to identify this.

\newcite{niven2019probing} show that high performance obtained from pre-trained language models such as BERT~\cite{devlin2018bert} are often achieved by exploiting spurious statistical cues in the dataset. We face a similar problem in our preliminary study when evaluating on a test set from a different domain. While the quantitative results of our models trained on Wikipedia data are extremely high, we observe a huge drop when testing on out-of-domain data. This motivates us to mitigate statistical cues in our data.

 Inspired by \textit{back-translation}~\citep{BT}, we generate paraphrases of our training data by introducing a pivot language and then translating the sentences back. This retains the semantics of the statements while removing existing stylistic biases. We back-translated both original Wikipedia sentences (i.e., positive samples) and the fabricated ones (i.e., negative samples). 
 Part C and D of Figure \ref{fig::training_data} show the back-translated versions of our positive and negative samples respectively.

\subsection{Inference Process}
\label{subsec:inference}

The framework of our inference stage is similar to the training procedure 
except that we also test on out-of-domain data. For each claim $s_i$ in test data, three model predictions $\{P^{en}(s^{en_i}), P^{cn}(s^{cn_i}),  P^{jp}(s^{jp_i})\}$ are generated. We then compute the colingual perspective difference of $s_i$ based on Equation \ref{eq:model}. Finally, we compute the correlation between model-predicted scores and human annotations. 


\begin{table}[t]
\small
\centering
\begin{tabular}{ccc}
\toprule
English (EN) & Chinese (CN) & Japanese (JP)\\ \midrule
0.53    & 0.61    & 0.58     \\ \bottomrule
\end{tabular}
\caption{Inter-rater agreement for English, Chinese and Japanese annotators using Krippendorff's alpha, with p-value \textless 1e-10. All annotators show moderate agreement within their own group.}
\label{table:inter rater}
\end{table}

\begin{table}[!t]
\small
\centering
\begin{tabular}{l|l|cc}
\toprule
\multicolumn{2}{c|}{}                 & \begin{tabular}[c]{@{}c@{}}Pearson\\ correlation\end{tabular} & \begin{tabular}[c]{@{}c@{}}Spearman\\ correlation\end{tabular} \\ \midrule
\multicolumn{2}{l|}{English-Chinese (E-C)}  & 0.26 (3e-3)                                                   & 0.27(2e-3)                                                     \\ 
\multicolumn{2}{l|}{Chinese-Japanese (C-J)} & 0.49(5e-9)                                                    & 0.50(2e-9)                                                     \\ 
\multicolumn{2}{l|}{Japanese-English (J-E)} & 0.29(7e-4)                                                    & 0.30(6e-4)                                                     \\ \bottomrule
\end{tabular}
\caption{Cross-group rater agreement, in terms of \textit{corr (p-value)}. We measure the correlation between collective judgements on 128 claims by raters from each pair of the colingual groups: \{E-C, C-J, J-E\}. }
\label{table:cross-rater}
\vspace{-4mm}
\end{table}

\begin{table*}[]
\small
\centering
\begin{tabular}{l|p{1.5cm}|p{1.6cm}|p{1.2cm}|l|p{1.3cm}|l}
\toprule
\multirow{2}{*}{\begin{tabular}[c]{@{}l@{}}\backslashbox{Training}{Testing}\end{tabular}} & \multicolumn{2}{l|}{No back-translation} & \multicolumn{2}{l|}{\begin{tabular}[c]{@{}l@{}}Translate only  negative\end{tabular}} & \multicolumn{2}{l}{Back-translate both} \\ \cline{2-7} 
                                                                            & Negative           & Positive           & Negative                                  & Positive                                  & Negative           & Positive           \\ \midrule
No back-translation                                                                       & 85.15          & 88.74          & 65.23                                 & 72.55                                 & 67.82          & 64.61          \\ 
Translate only negative                                                                    & 79.78          & 87.11          & 92.06                                & 94.53                                & 77.26          & 76.31          \\ 
Back-translate both                                                                        & \textbf{92.56} & \textbf{94.88} & \textbf{92.17}                        & \textbf{95.69}                        & \textbf{87.10} & \textbf{91.92} \\ \bottomrule
\end{tabular}
\caption{F1 scores of positive and negative class respectively, with models trained under three different settings: 1) neither the positive or the negative samples are back-translated, 2) only negative samples are back-translated, and 3) both positive and negative samples are back-translated. We then test them on the same held-out dataset.}
\vspace*{-3mm}
\label{table:bt}
\end{table*}


\label{sec:exp}

\section{Experiments}

\subsection{Experimental Setup}\label{subsec:exp_set}
For all classifiers, we start the sentence representations with BERT-base \cite{devlin2018bert} model, and then fine-tune them during training. We set sequence length as 128, batch size as 64 and learning rate as $2\mathrm{e}{-5}$. We also study the efficiency of back-translation on reducing stylistic biases.
Specifically, we train BERT models using data from 3 different settings: 1) no back-translation, 
2) back-translate only negative samples, and
3) back-translate both positive and negative samples.

\subsection{Binarization}

We binarize the ground truth (with 0.5 as threshold) for the simplicity of data collection. Here 0 represents that a colingual group tends to maintain an opposite perspective, while 1 indicates a group tends to agree with the claim. For Wikipedia sentences, which we use for training and in-domain evaluation, the sentences originally selected from Wikipedia are positive (1) while the one we modified algorithmically are negative (0).

\subsection{Inter and cross-group rater agreement}
To show how the annotators within a colingual group agree with each other, we calculate the inter-annotator agreement (IAA) using Krippendorff's alpha. We also leverage attention questions to remove irresponsible annotators. The final IAAs are listed in Table \ref{table:inter rater}. For all three languages, the correlation within a culture is above 0.5, demonstrating that the annotators are moderately correlated.

We also investigate how cross-group raters agree with each other, and calculate their Pearson and Spearman correlation (as listed in Table \ref{table:cross-rater}). The Chinese and Japanese raters have higher correlation with each other than they are with English raters.

\subsection{Baselines}
We compare our proposed Colingual Perspective Identifier (CLUSTER) with these baselines:

    \textbf{Random}: Random numbers within $[0,1]$ are generated to simulate model predictions of all perspective classifiers.
    
    \textbf{LM}: We regard the average of word-level log probability (sentence log probability divided by length) generated by multi-lingual GPT2 \cite{radford2019language, GPT2-ML, GPT2-Japanese} as model predictions. We then use the min-max method to normalize the log probabilities.
    
    \textbf{Weak CLUSTER}: Our proposed Colingual Perspective Identifier, trained on Wikipedia sentences \textit{without} \textit{back-translation}.

\section{Results}
\label{sec:results}

\subsection{The Effects of Back-translation}
Table \ref{table:bt} shows that models trained with \textit{no back-translation} and \textit{translate only negative} work well under their own respective setting, but does not transfer well to other scenarios. On the other hand, we obtain best and most robust results when the model is trained on data being back-translated for both positive and negative samples. Hence, back-translation (for both positive and negative samples) is ideal to be used for inference in other domains.

\subsection{Agreement between Model Prediction and Human Annotation }

\begin{table*}[ht!]
\centering
\small
\begin{tabular}{c|c|c|c|c|c|c|c}
\toprule
Model                                                                  & \begin{tabular}[c]{@{}c@{}}Correlation\\ Type\end{tabular} & \begin{tabular}[c]{@{}c@{}}English\\ (EN)\end{tabular} & \begin{tabular}[c]{@{}c@{}}Chinese\\ (CN)\end{tabular} & \begin{tabular}[c]{@{}c@{}}Japanese\\ (JP)\end{tabular} & \begin{tabular}[c]{@{}c@{}}Cross-culture\\ (E-C)\end{tabular} & \begin{tabular}[c]{@{}c@{}}Cross-culture\\ (C-J)\end{tabular} & \begin{tabular}[c]{@{}c@{}}Cross-culture\\ (J-E)\end{tabular} \\ \midrule
                                                                       & Pearson                                                    & 0.00(0.5)                                              & 0.00(0.5)                                              & 0.00(0.5)                                               & 0.00(0.5)                                                     & 0.00(0.5)                                                     & 0.00(0.5)                                                     \\  
\multirow{-2}{*}{Random}                                               & Spearman                                                   & 0.00(0.5)                                              & 0.00(0.5)                                              & 0.00(0.5)                                               & 0.00(0.5)                                                     & 0.00(0.5)                                                     & 0.00(0.5)                                                     \\ \hline
                                                                       & Pearson                                                    & 0.17 (0.05)                                            & 0.07 (0.42)                                            & 0.12(0.19)                                              & 0.11 (0.23)                                                   & 0.08(0.36)                                                    & 0.15(0.09)                                                    \\  
\multirow{-2}{*}{LM}                                                   & Spearman                                                   & 0.16 (0.08)                                            & 0.08 (0.35)                                            & 0.11(0.22)                                              & 0.09 (0.30)                                                   & 0.09(0.33)                                                    & 0.13(0.14)                                                    \\ \hline
                                                                       & Pearson                                                    & 0.22 (0.01)                                            & 0.19 (0.03)                                            & 0.18(0.05)                                              & 0.03 (0.73)                                                   & 0.05(0.61)                                                    & 0.15(0.09)                                                    \\  
\multirow{-2}{*}{\begin{tabular}[c]{@{}c@{}}Weak \\ CLUSTER\end{tabular}} & Spearman                                                   & 0.11 (0.23)                                            & 0.13 (0.14)                                            & 0.10(0.28)                                              & 0.07 (0.42)                                                   & 0.06(0.51)                                                    & 0.11(0.23)                                                    \\ \hline
{\color[HTML]{C00000} }                                                & {\color[HTML]{C00000} Pearson}                             & {\color[HTML]{C00000} \textbf{0.37 (1e-5)}}            & {\color[HTML]{C00000} \textbf{0.41 (1e-6)}}            & {\color[HTML]{C00000} \textbf{0.40(3e-6)}}              & {\color[HTML]{C00000} \textbf{0.25 (4e-3)}}                   & {\color[HTML]{C00000} \textbf{0.20(0.02)}}                    & {\color[HTML]{C00000} \textbf{0.35(4e-5)}}                    \\ 
\multirow{-2}{*}{{\color[HTML]{C00000} CLUSTER}}                          & Spearman                                                   & 0.32 (2e-4)                                            & 0.34 (5e-4)                                            & 0.39(6e-6)                                              & 0.21 (0.01)                                                   & 0.18(0.04)                                                    & 0.31(4e-4)                                                    \\ \bottomrule
\end{tabular}

\caption{Agreement between model predictions and human annotations, in the format of \textit{correlation (p-value)}. A higher value on Pearson correlation over Spearman correlation indicates that linear correlation is more significant than the rank correlation, and vice versa.}
\label{table:model_correlation}
\vspace{-1em}
\end{table*}

Table \ref{table:model_correlation} reports the the correlations between the CLUSTER and baseline models with human annotations. We observe that the Random method does not capture any perspective representations at all. A competitive language model such as GPT-2 can bring significant improvements over Random because it is trained on a very large NLP corpus (including English Wikipedia), where group perspectives are implicitly included. 

Moreover, the performance of Weak CLUSTER is partially better then language models, but still rather limited, probably due to style bias in negative samples. Finally, we can find that CLUSTER consistently outperforms all its competitors, and obtains 0.10 $\sim$ 0.22 performance gains over the second best model for all three colingual groups. 

Last, we want to point out that unlike many other NLP tasks, the IAA (or human performance) should not be viewed as golden or an upper-bound in our evaluation. The IAA is just an indicator of how unanimous the annotators are on diverse concepts, including very controversial topics such as abortion. Therefore, machine-human correlation can reasonably be higher than within-human correlation.

\begin{table}[!t]
\small
\centering
\begin{tabular}{@{ }c@{ }|@{ }c@{ }@{ }c@{ }@{ }c@{ }|@{ }c@{ }@{ }c@{ }@{ }c@{ }}
\toprule
Model                                               & EN                          & CN                          & JP                          & E-C                       & C-J                       & J-E                       \\ \midrule
Random                                              & 0.50                        & 0.50                        & 0.50                        & 0.50                        & 0.50                        & 0.50                        \\ 
LM                                                  & 0.60                        & 0.50                        & 0.54                        & 0.53                        & 0.55                        & 0.56                        \\ 
Weak CLUSTER& 0.70                        & 0.55                        & 0.56                        & 0.45                        & 0.53                        & 0.52                        \\ 
{\color[HTML]{C00000} CLUSTER}                         & {\color[HTML]{C00000} \textbf{0.73}} & {\color[HTML]{C00000} \textbf{0.64}} & {\color[HTML]{C00000} \textbf{0.66}} & {\color[HTML]{C00000} \textbf{0.58}} & {\color[HTML]{C00000} \textbf{0.60}} & {\color[HTML]{C00000} \textbf{0.63}} \\ 
\bottomrule
\end{tabular}
\vspace*{-1mm}
\caption{The binary accuracy. We test both within \{EN, CN, JP\} and across \{E-C, C-J, J-E\} groups. For scores within a culture($\in [0,1]$), the threshold is set to 0.5. For cross-group perspective scores ($\in [-1,1]$), the threshold is set to 0.}
\vspace*{-5mm}
\label{table:binary accuracy}
\end{table}

\subsection{Binary Accuracy}
To further investigate the performance of our model and the baselines, we calculate the number of instances where binarized predictions and ground truths match with each other. The results are shown in Table \ref{table:binary accuracy}. Again, our CLUSTER model achieved the best performance in all aspects.




\section{Qualitative Analysis}
\label{sec:quanlitative_analysis}


While section \ref{sec:results} shows quantitative results and correlation values, we want to understand the advantages of our model on a qualitative basis. To this end, we select 50 claims from five particular topics: \textit{marriage}, \textit{corruption}, \textit{cuisine}, \textit{christmas} and \textit{baseball}, and then obtain CLUSTER model predictions on these claims. We do not collect human annotations for these sentences, but use them only for qualitative analysis and visualization purposes detailed below. 

\begin{figure*}[!t]
\centering
\includegraphics[width = 1.0\linewidth]{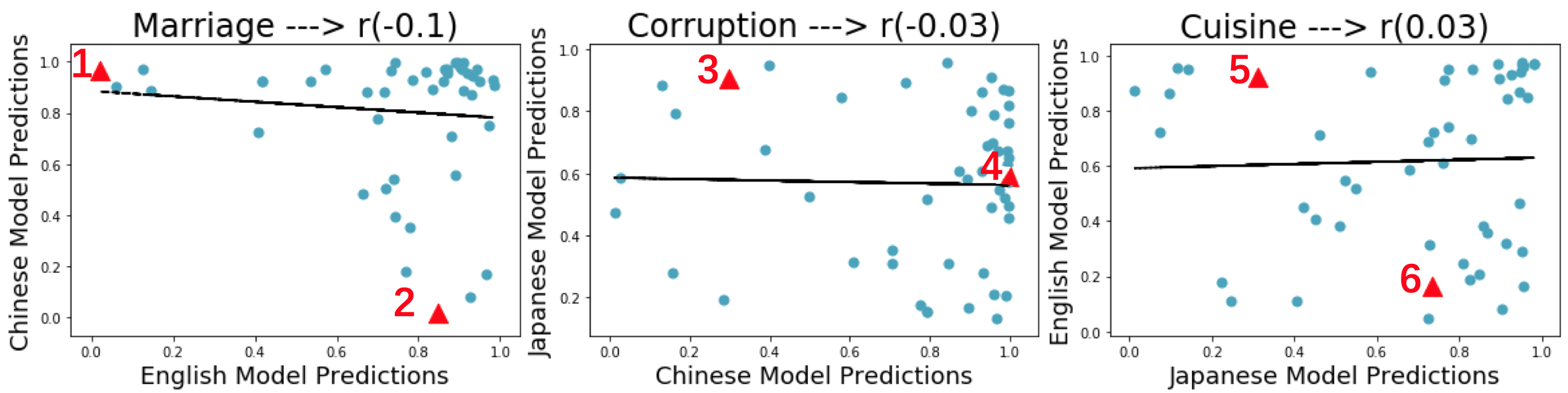}
\vspace{-3mm}
\caption{\label{figure:pie} Model predictions on  crosslingually disagreed topics: \textit{marriage}, \textit{corruption} and \textit{cuisine} with their correlation values. Each dot (or triangle) represents one of the claims randomly selected from IMHO, with the x-y axis representing \{E-C, C-J, J-E\} predicted scores in sequence. Red triangular points are the following sentences: \textbf{1.} \textit{Marriage is not about meeting someone you connect to, but both people being matured, and in the same headspace.} \textbf{2.} \textit{If he cannot share his concerns with her, he is poor marriage material.} \textbf{3.} \textit{If you don't reveal others' corruption you are culpable as well.} \textbf{4.} \textit{There is plenty of corruption pulled out in the open these days, and that has been happening at a faster pace than ever before.} \textbf{5.} \textit{Mexican, Mediterranean, Indian and Thai cuisines have the most delicious vegetarian dishes.} \textbf{6.} \textit{Grilled fish is much better cooked at home and shared with friends.} }
\vspace{-2mm}
\end{figure*}

\begin{figure}[!tp] 
	\setlength{\abovecaptionskip}{0.1cm}
	\setlength{\belowcaptionskip}{-0.5cm}
    \centering
    \includegraphics[width = 1.0\linewidth]{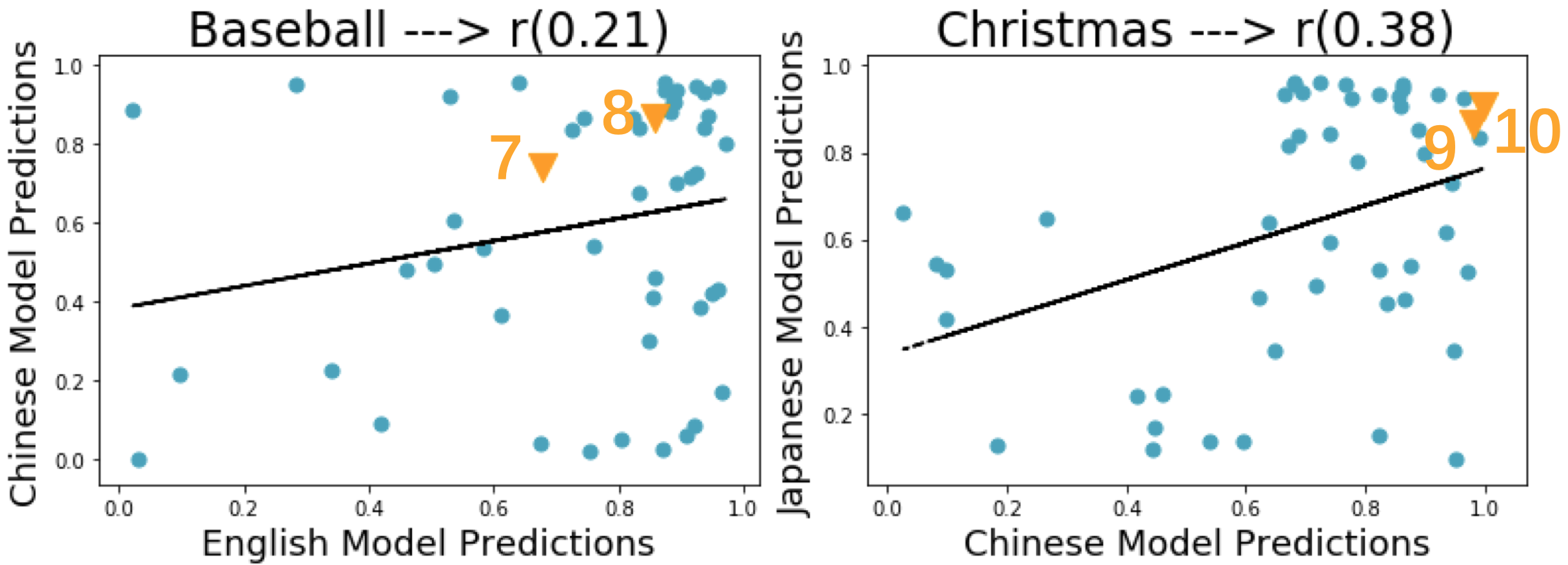}
    \caption{Model predictions on cross-lingually agreed topics: \textit{baseball} and \textit{christmas}, along with their correlation. The meaning of dots is the same as Figure \ref{figure:pie}. Orange triangles represent the following sentences: \textbf{7.} \textit{Cricket is as fun to play as baseball if you limit the ``innings” or overs.} \textbf{8.} \textit{Things like basketball, baseball, tennis, golf, etc. are far more popular globally.} \textbf{9.} \textit{Christmas, even minus the religious meanings, has good attributes in theory but has been too commercialized. } \textbf{10.} \textit{
   I believe in giving gifts to kids because, Christmas is for children. } }
    \label{fig:agreement}
    \vspace{-1mm}
\end{figure}

For each colingual group pair in \{E-C, C-J, J-E\} and a given topic, we report the visualization of 50 claim pairs in Figure \ref{figure:pie} and \ref{fig:agreement}. Here, each dot (or triangle) represents one of the 50 claims which are randomly selected from IMHO, with the x-y axis representing the \{E-C, C-J, J-E\} model predictions. The blue dots that fall along the diagonals are where the two models agree. On the contrary, dots that fall on the upper left or the lower right part are where the models do not agree with each other. For example, sentence 1  in Figure \ref{figure:pie} is closer to the Chinese culture (upper left corner), while  English speakers tend to agree more with sentence 2 (lower right corner). We select representative examples in each region and list them in the captions.

 First, from Figure \ref{figure:pie} we observe that the model pairs have zero or negative correlation on three topics: \textit{marriage}, \textit{corruption} and \textit{cuisine}, suggesting that the corresponding language speakers take contrasting stances towards these topics. Second, Figure \ref{fig:agreement} shows that 1) the English and Chinese speakers hold similar views on \textit{baseball}, and 2) the Chinese and Japanese speakers share similar views on \textit{christmas}. For example, Christmas, which is not a traditional holiday in East Asia, is adopted directly from the western world. The Chinese and Japanese speakers both follow the western customs and hence view Christmas likewise.

\section{Related Work}
\label{sec:related}

\paragraph{Online Disagreement}
Most works about online disagreement focus on a single culture or language \cite{sridhar-etal-2015-joint, wang-yang-2015-thats,sridhar-etal-2015-joint,rosenthal-mckeown-2015-couldnt}, thus are restricted to a single group. While these works try to computationally model disagreement or stance in debates, they do not target at finding cultural or cross-group differences. We, on the other hand, aim at understanding the disagreement in perspectives through different colingual groups according to their respective languages.

\paragraph{Cultural Study in Blogs or Social Media}
\newcite{nakasaki2009visualizing} present a framework to visualize the cross-cultural differences in multilingual blogs. \newcite{elahi-monachesi-2012-examination} show that using emotion terms as culture features is effective in analyzing cross-cultural difference in social media data. However, it is only restricted to a single topic (love and relationship). In contrast, we use Wikipedia to study cross-group differences in perspectives on a much larger scale and do not restrict ourselves to one single topic.

\paragraph{Cultural Difference in Word Usage}
\newcite{garimella2018quantifying} investigate the cross-cultural differences in word usages between Australian and American English through \textit{socio-linguistic features} supervisedly. \newcite{garimella-etal-2016-identifying} use
social network structures and user interactions, to study how to
quantify the controversy of topics within a culture
and language. \newcite{gutierrez2016detecting} detect differences of word usage in the cross-lingual topics of multilingual topic modeling results. \newcite{lin-etal-2018-mining} present distributional approaches to compute cross-cultural differences or similarities between two terms from
different cultures focusing primarily on named entities. Our work is not limited to word usage or any particular topics. Instead, we focus on understanding cross-group differences of perspective at the sentence level. 

\paragraph{Argumentation}
In argumentation, \textit{Framing} is used to emphasize a specific aspect of a controversial topic. \newcite{ajjour-etal-2019-modeling} introduce frame identification, which is the task of splitting a set of arguments into non-overlapping frames. \newcite{chen2019seeing} also release a dataset of claims, perspectives and evidence and propose the task of substantiated perspective discovery where, given a claim, a system is expected to discover a diverse set of well-corroborated perspectives that take a stance with respect to the claim. Different interests, cultural and cultural backgrounds diverge people from on taking a certain course of action. While both works deal with different perspectives about arguments in English, our work focuses on identifying the differences from a cross-lingual point of view. 
\section{Conclusion}
\label{sec:conclusion}

We present CLUSTER, a computational method to identify distributional differences in cross-group perspectives, and evaluate it with human judgements. Through detailed experiments, we show that CLUSTER is straightforward and effective. Furthermore, we show CLUSTER generalizes well for out-of-domain scenarios by training the group perspective models on Wikipedia and test on claims collected from Reddit. This means that the proposed method learns the task, not the data. Besides, the general model of perspective difference identification can be useful in many NLP tasks such as fact checking, sentiment analysis, as well as cross-cultural studies in computational social science or multilingual debate forums.

 As a first attempt towards automatic identification of cross-cultural differences, our work still has much room for improvement. Future directions include more complicated ways of composing negative samples, more well-crafted models, and extending our pipeline to fine-grained subgroups speaking the same language, especially for English as a global language spoken by many nations.


\bibliography{anthology,emnlp2020}

\begin{thebibliography}{27}
\expandafter\ifx\csname natexlab\endcsname\relax\def\natexlab#1{#1}\fi

\bibitem[{Ajjour et~al.(2019)Ajjour, Alshomary, Wachsmuth, and
  Stein}]{ajjour-etal-2019-modeling}
Yamen Ajjour, Milad Alshomary, Henning Wachsmuth, and Benno Stein. 2019.
\newblock Modeling frames in argumentation.
\newblock In \emph{Proceedings of the 2019 Conference on Empirical Methods in
  Natural Language Processing and the 9th International Joint Conference on
  Natural Language Processing (EMNLP-IJCNLP)}, pages 2922--2932.

\bibitem[{Bar-Tal(2000)}]{bar2000shared}
Daniel Bar-Tal. 2000.
\newblock \emph{Shared beliefs in a society: Social psychological analysis}.
\newblock Sage Publications.

\bibitem[{Bracewell and Tomlinson(2012)}]{bracewell2012language}
David Bracewell and Marc Tomlinson. 2012.
\newblock The language of power and its cultural influence.
\newblock In \emph{Proceedings of {COLING} 2012: Posters}.

\bibitem[{Callahan and Herring(2011)}]{callahan2011cultural}
Ewa~S Callahan and Susan~C Herring. 2011.
\newblock Cultural bias in wikipedia content on famous persons.
\newblock \emph{Journal of the American society for information science and
  technology}, 62(10):1899--1915.

\bibitem[{Chakrabarty et~al.(2019)Chakrabarty, Hidey, and
  McKeown}]{chakrabarty2019imho}
Tuhin Chakrabarty, Christopher Hidey, and Kathleen McKeown. 2019.
\newblock Imho fine-tuning improves claim detection.
\newblock \emph{arXiv preprint arXiv:1905.07000}.

\bibitem[{Chen et~al.(2019)Chen, Khashabi, Yin, Callison-Burch, and
  Roth}]{chen2019seeing}
Sihao Chen, Daniel Khashabi, Wenpeng Yin, Chris Callison-Burch, and Dan Roth.
  2019.
\newblock Seeing things from a different angle: Discovering diverse
  perspectives about claims.
\newblock \emph{arXiv preprint arXiv:1906.03538}.

\bibitem[{Devlin et~al.(2018)Devlin, Chang, Lee, and
  Toutanova}]{devlin2018bert}
Jacob Devlin, Ming-Wei Chang, Kenton Lee, and Kristina Toutanova. 2018.
\newblock Bert: Pre-training of deep bidirectional transformers for language
  understanding.
\newblock \emph{arXiv preprint arXiv:1810.04805}.

\bibitem[{Elahi and Monachesi(2012)}]{elahi-monachesi-2012-examination}
Mohammad~Fazleh Elahi and Paola Monachesi. 2012.
\newblock \href
  {http://www.lrec-conf.org/proceedings/lrec2012/pdf/942_Paper.pdf} {An
  examination of cross-cultural similarities and differences from social media
  data with respect to language use}.
\newblock In \emph{Proceedings of the Eighth International Conference on
  Language Resources and Evaluation ({LREC}-2012)}, pages 4080--4086, Istanbul,
  Turkey. European Languages Resources Association (ELRA).

\bibitem[{Garimella et~al.(2016)Garimella, Mihalcea, and
  Pennebaker}]{garimella-etal-2016-identifying}
Aparna Garimella, Rada Mihalcea, and James Pennebaker. 2016.
\newblock \href {https://www.aclweb.org/anthology/C16-1065} {Identifying
  cross-cultural differences in word usage}.
\newblock In \emph{Proceedings of {COLING} 2016, the 26th International
  Conference on Computational Linguistics: Technical Papers}, pages 674--683,
  Osaka, Japan. The COLING 2016 Organizing Committee.

\bibitem[{Garimella et~al.(2018)Garimella, Morales, Gionis, and
  Mathioudakis}]{garimella2018quantifying}
Kiran Garimella, Gianmarco De~Francisci Morales, Aristides Gionis, and Michael
  Mathioudakis. 2018.
\newblock Quantifying controversy on social media.
\newblock \emph{ACM Transactions on Social Computing}, 1(1):3.

\bibitem[{Gelman and Roberts(2017)}]{gelman2017language}
Susan~A Gelman and Steven~O Roberts. 2017.
\newblock How language shapes the cultural inheritance of categories.
\newblock \emph{Proceedings of the National Academy of Sciences},
  114(30):7900--7907.

\bibitem[{Guti{\'e}rrez et~al.(2016)Guti{\'e}rrez, Shutova, Lichtenstein,
  de~Melo, and Gilardi}]{gutierrez2016detecting}
E~Dario Guti{\'e}rrez, Ekaterina Shutova, Patricia Lichtenstein, Gerard
  de~Melo, and Luca Gilardi. 2016.
\newblock Detecting cross-cultural differences using a multilingual topic
  model.
\newblock \emph{Transactions of the Association for Computational Linguistics},
  4:47--60.

\bibitem[{Hasan and Ng(2014)}]{hasan-ng-2014-taking}
Kazi~Saidul Hasan and Vincent Ng. 2014.
\newblock \href {https://doi.org/10.3115/v1/D14-1083} {Why are you taking this
  stance? identifying and classifying reasons in ideological debates}.
\newblock In \emph{Proceedings of the 2014 Conference on Empirical Methods in
  Natural Language Processing ({EMNLP})}, pages 751--762, Doha, Qatar.
  Association for Computational Linguistics.

\bibitem[{Heckman(1977)}]{heckman1977sample}
James~J Heckman. 1977.
\newblock Sample selection bias as a specification error (with an application
  to the estimation of labor supply functions).
\newblock Technical report, National Bureau of Economic Research.

\bibitem[{Hoang et~al.(2018)Hoang, Koehn, Haffari, and Cohn}]{BT}
Vu~Cong~Duy Hoang, Philipp Koehn, Gholamreza Haffari, and Trevor Cohn. 2018.
\newblock \href {https://doi.org/10.18653/v1/W18-2703} {Iterative
  back-translation for neural machine translation}.
\newblock In \emph{Proceedings of the 2nd Workshop on Neural Machine
  Translation and Generation}, pages 18--24, Melbourne, Australia. Association
  for Computational Linguistics.

\bibitem[{Khaslavsky(1998)}]{khaslavsky1998integrating}
Julie Khaslavsky. 1998.
\newblock Integrating culture into interface design.
\newblock In \emph{CHI 98 conference summary on Human factors in computing
  systems}, pages 365--366. ACM.

\bibitem[{Lin et~al.(2018)Lin, Xu, Zhu, and Hwang}]{lin-etal-2018-mining}
Bill~Yuchen Lin, Frank~F. Xu, Kenny Zhu, and Seung-won Hwang. 2018.
\newblock \href {https://doi.org/10.18653/v1/P18-1066} {Mining cross-cultural
  differences and similarities in social media}.
\newblock In \emph{Proceedings of the 56th Annual Meeting of the Association
  for Computational Linguistics (Volume 1: Long Papers)}, pages 709--719,
  Melbourne, Australia. Association for Computational Linguistics.

\bibitem[{Nakasaki et~al.(2009)Nakasaki, Kawaba, Yamazaki, Utsuro, and
  Fukuhara}]{nakasaki2009visualizing}
Hiroyuki Nakasaki, Mariko Kawaba, Sayuri Yamazaki, Takehito Utsuro, and
  Tomohiro Fukuhara. 2009.
\newblock Visualizing cross-lingual/cross-cultural differences in concerns in
  multilingual blogs.
\newblock In \emph{Third International AAAI Conference on Weblogs and Social
  Media}.

\bibitem[{Ng et~al.(2019)Ng, Yee, Baevski, Ott, Auli, and
  Edunov}]{ng2019facebook}
Nathan Ng, Kyra Yee, Alexei Baevski, Myle Ott, Michael Auli, and Sergey Edunov.
  2019.
\newblock Facebook fair's wmt19 news translation task submission.
\newblock \emph{arXiv preprint arXiv:1907.06616}.

\bibitem[{Niven and Kao(2019)}]{niven2019probing}
Timothy Niven and Hung-Yu Kao. 2019.
\newblock Probing neural network comprehension of natural language arguments.
\newblock \emph{arXiv preprint arXiv:1907.07355}.

\bibitem[{Radford et~al.(2019)Radford, Wu, Child, Luan, Amodei, and
  Sutskever}]{radford2019language}
Alec Radford, Jeff Wu, Rewon Child, David Luan, Dario Amodei, and Ilya
  Sutskever. 2019.
\newblock Language models are unsupervised multitask learners.

\bibitem[{Rosenthal and McKeown(2015)}]{rosenthal-mckeown-2015-couldnt}
Sara Rosenthal and Kathy McKeown. 2015.
\newblock \href {https://doi.org/10.18653/v1/W15-4625} {{I} couldn{'}t agree
  more: The role of conversational structure in agreement and disagreement
  detection in online discussions}.
\newblock In \emph{Proceedings of the 16th Annual Meeting of the Special
  Interest Group on Discourse and Dialogue}, pages 168--177, Prague, Czech
  Republic. Association for Computational Linguistics.

\bibitem[{Sakamoto(2019)}]{GPT2-Japanese}
Toshiyuki Sakamoto. 2019.
\newblock Japanese gpt2 generation model.
\newblock \url{https://github.com/tanreinama/gpt2-japanese}.

\bibitem[{Sperber and Hirschfeld(2004)}]{sperber2004foundations}
Dan Sperber and Lawrence~A Hirschfeld. 2004.
\newblock The cognitive foundations of cultural stability and diversity.
\newblock \emph{Trends in cognitive sciences}, 8(1):40--46.

\bibitem[{Sridhar et~al.(2015)Sridhar, Foulds, Huang, Getoor, and
  Walker}]{sridhar-etal-2015-joint}
Dhanya Sridhar, James Foulds, Bert Huang, Lise Getoor, and Marilyn Walker.
  2015.
\newblock \href {https://doi.org/10.3115/v1/P15-1012} {Joint models of
  disagreement and stance in online debate}.
\newblock In \emph{Proceedings of the 53rd Annual Meeting of the Association
  for Computational Linguistics and the 7th International Joint Conference on
  Natural Language Processing (Volume 1: Long Papers)}, pages 116--125,
  Beijing, China. Association for Computational Linguistics.

\bibitem[{Wang and Yang(2015)}]{wang-yang-2015-thats}
William~Yang Wang and Diyi Yang. 2015.
\newblock \href {https://doi.org/10.18653/v1/D15-1306} {That{'}s so
  annoying!!!: A lexical and frame-semantic embedding based data augmentation
  approach to automatic categorization of annoying behaviors using {\#}petpeeve
  tweets}.
\newblock In \emph{Proceedings of the 2015 Conference on Empirical Methods in
  Natural Language Processing}, pages 2557--2563, Lisbon, Portugal. Association
  for Computational Linguistics.

\bibitem[{Zhang(2019)}]{GPT2-ML}
Zhibo Zhang. 2019.
\newblock Gpt2-ml: Gpt-2 for multiple languages.
\newblock \url{https://github.com/imcaspar/gpt2-ml}.

\end{thebibliography}
\bibliographystyle{acl_natbib}

\clearpage
\appendix
\paragraph{Appendix}
\section{Hyper-parameters and other Experimental Settings}
To train the classifiers, we start the sentence representations with the pre-trained BERT \cite{devlin2018bert} model, and then fine-tune them. For all models, we set sequence length as 128, batch size as 64 and learning rate as $2\mathrm{e}{-5}$. We train each CLUSTER model for 5 epochs and save the best model only.

\begin{enumerate}
    \item \textbf{Number of parameters:}\quad Each CLUSTER model, fine-tuned on the BERT-base model, has 102M trainable parameters. 
    \item \textbf{Runtime:} \quad Our average training time is 2 to 5 hours, depending on the size of training data for each language (see Table \ref{table:data}).
    \item \textbf{Hardware configuration:}\quad  We use three GeForce RTX 2080 GPUs.
    \item \textbf{Hyper-parameter tuning:}\quad We manually tune the hyper-parameters and report the configuration that has the best F1 score on our validation set.  
\end{enumerate}

\section{Topics and Visualization}
The sixteen topics that are selected for evaluation, along with the Pearson correlations of culture model predictions on 50 randomly sampled sentences, are listed in Table \ref{table:16topics}. We highlight the topics with relatively high and low values of correlation coefficients in red and blue. Note that we do not collect human annotations for these sentences, but use them only for qualitative analysis and visualization purposes. 

\begin{figure}[]
\centering
\includegraphics[width = 0.75\linewidth]{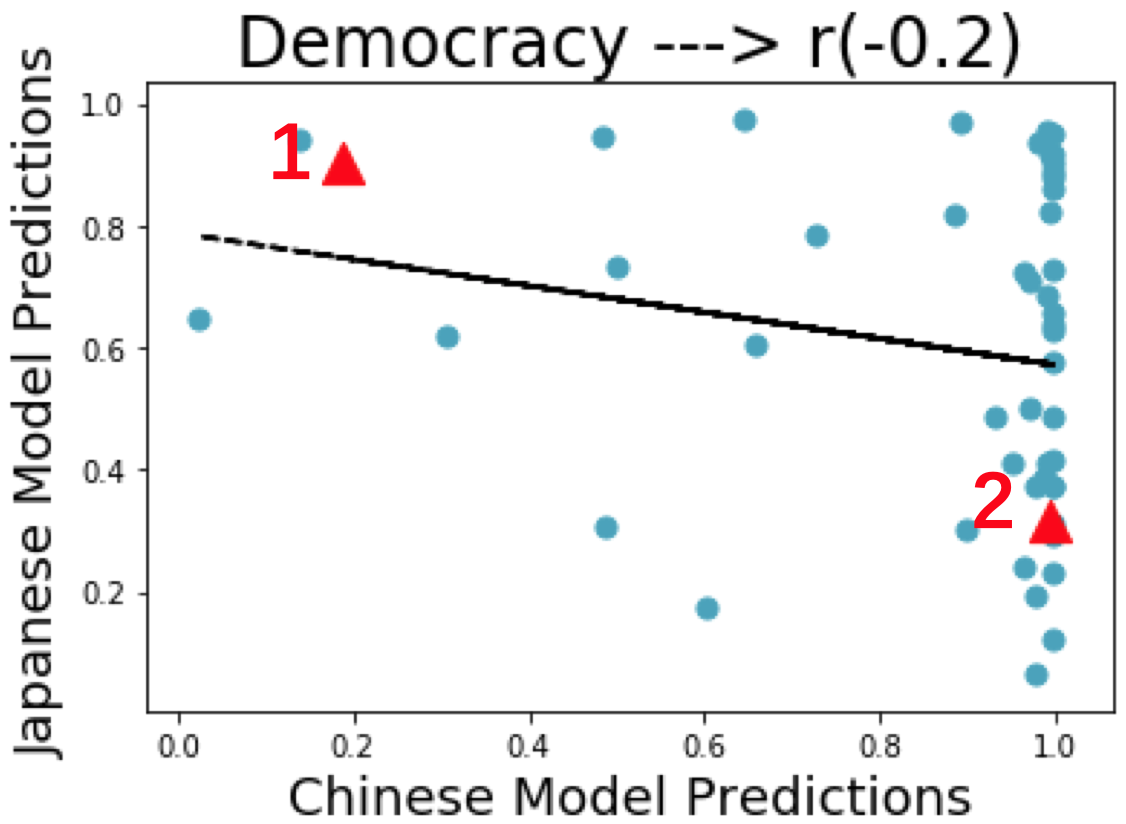}
\caption{Model predictions on \textit{democracy} of Chinese (x-axis) and Japanese (y-axis) models, and the correlation coefficient. The red triangles represent cross-lingually disagreed sentences: \textbf{1.} \textit{Yeah, mandatory voting should be a required part of a democracy.} \textbf{2.} \textit{The ideal system would be a merger of democracy and socialism (which we  are slowly moving towards).}}
\label{fig:democracy}
\end{figure}

 As can be seen, most topics have a positive correlation, meaning that the English, Chinese and Japanese colingual groups have a general agreements on most subjects such as 
 \textit{savings}, \textit{baseball} and \textit{cheese}. Christmas, which is not a traditional holiday in China or Japan, is adopted directly from the western world. That's why all the three models view Christmas likewise. In addition, the models have dispute on topics such as \textit{bible}, \textit{marriage},  \textit{corruption}, and \textit{abortion}. To get a more intuitive sense of the score distribution, we further visualize the model-predicted scores on more topics in Figure \ref{fig:democracy} and Figure \ref{fig:savings}. 

\begin{table}[]
\small
\centering
\begin{tabular}{|c|l|l|l|}
\hline
      Topics      & \multicolumn{1}{c|}{E-C}            & \multicolumn{1}{c|}{C-J}            & \multicolumn{1}{c|}{J-E}            \\ \hline
Savings     & 0.09 (0.51)                         & {\color[HTML]{9A0000} 0.40 (0.00)}  & 0.31 (0.03)                         \\ \hline
Cuisine     & 0.01 (0.97)                         &  0.21 (0.14)  & 0.03 (0.83)                         \\ \hline
Christmas   & {\color[HTML]{9A0000} 0.37 (0.01)}  & {\color[HTML]{9A0000} 0.38 (0.01)}  & {\color[HTML]{9A0000} 0.45 (0.00)}  \\ \hline
Bible       & {\color[HTML]{00009B} -0.02 (0.89)} & 0.16 (0.26)                         & 0.16 (0.28)                         \\ \hline
Soup        & 0.26 (0.07)                         & 0.15 (0.30)                         & 0.22 (0.12)                         \\ \hline
Terrorism   & 0.09 (0.51)                         & 0.07 (0.61)                         & 0.20 (0.16)                         \\ \hline
Marriage    & {\color[HTML]{303498} -0.10 (0.50)} & 0.21 (0.13)                         & 0.04 (0.81)                         \\ \hline
Corruption  & 0.11 (0.44)                         & {\color[HTML]{303498} -0.04 (0.77)} & {\color[HTML]{303498} -0.09 (0.54)} \\ \hline
Baseball    & 0.21 (0.13)                         & 0.13 (0.35)                         & 0.11 (0.46)                         \\ \hline
Cheese      & 0.17 (0.23)                         & 0.03 (0.83)                         & 0.28 (0.05)                         \\ \hline
Communism   & 0.06 (0.67)                         & 0.03 (0.83)                         & 0.10 (0.48)                         \\ \hline
Democracy   & 0.09 (0.56)                         & {\color[HTML]{303498} -0.20 (0.16)} & 0.17 (0.23)                         \\ \hline
Russia      & 0.29 (0.04)                         & {\color[HTML]{9A0000} 0.33 (0.02)}  & 0.16 (0.27)                         \\ \hline
Abortion    & {\color[HTML]{303498} -0.04 (0.77)} & 0.07 (0.65)                         & {\color[HTML]{9A0000} 0.33 (0.02)}  \\ \hline
Racism      & {\color[HTML]{9A0000} 0.31 (0.03)}  & 0.18 (0.20)                         & 0.12 (0.40)                         \\ \hline
Gun control & 0.07 (0.63)                         & 0.15 (0.30)                         & 0.19 (0.18)                         \\ \hline
\end{tabular}
\caption{The sixteen topics that are selected for evaluation, along with the correlations between English-Chinese (E-C), Chinese-Japanese (C-J), and Japanese-English (J-E) culture model predictions on 50 randomly sampled sentences, in terms of \textit{corr (p-value)}. }
\label{table:16topics}
\end{table}

\begin{figure}[]
\centering
\includegraphics[width = 0.75\linewidth]{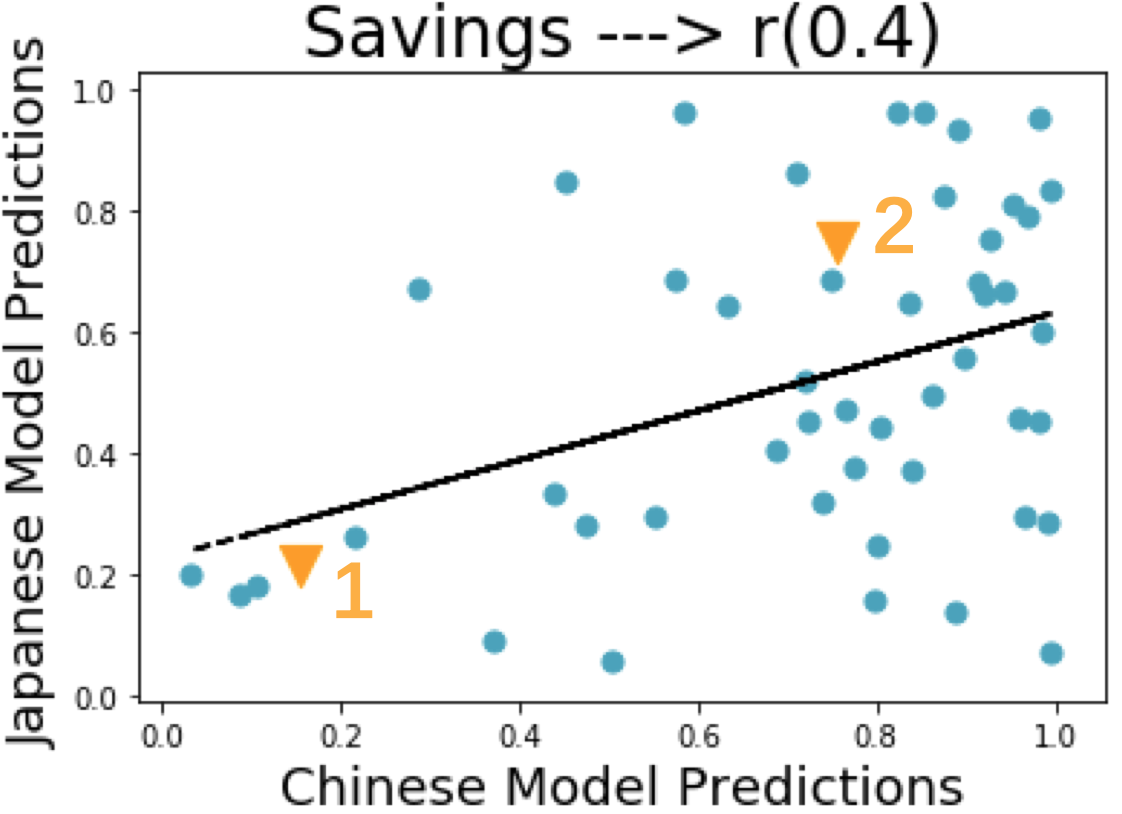}
\caption{Model predictions on \textit{savings} of Chinese (x-axis) and Japanese (y-axis) models, and the correlation coefficient. The orange triangles represent cross-lingually agreed sentences: \textbf{1.} \textit{Higher risk-free interest is needed to stimulate savings and to avoid credit recessions.} \textbf{2.} \textit{Life savings essentially means to me what you are gonna leave to your heirs.}}
\label{fig:savings}
\end{figure}

\begin{figure}[]
\centering
\includegraphics[width = 0.75\linewidth]{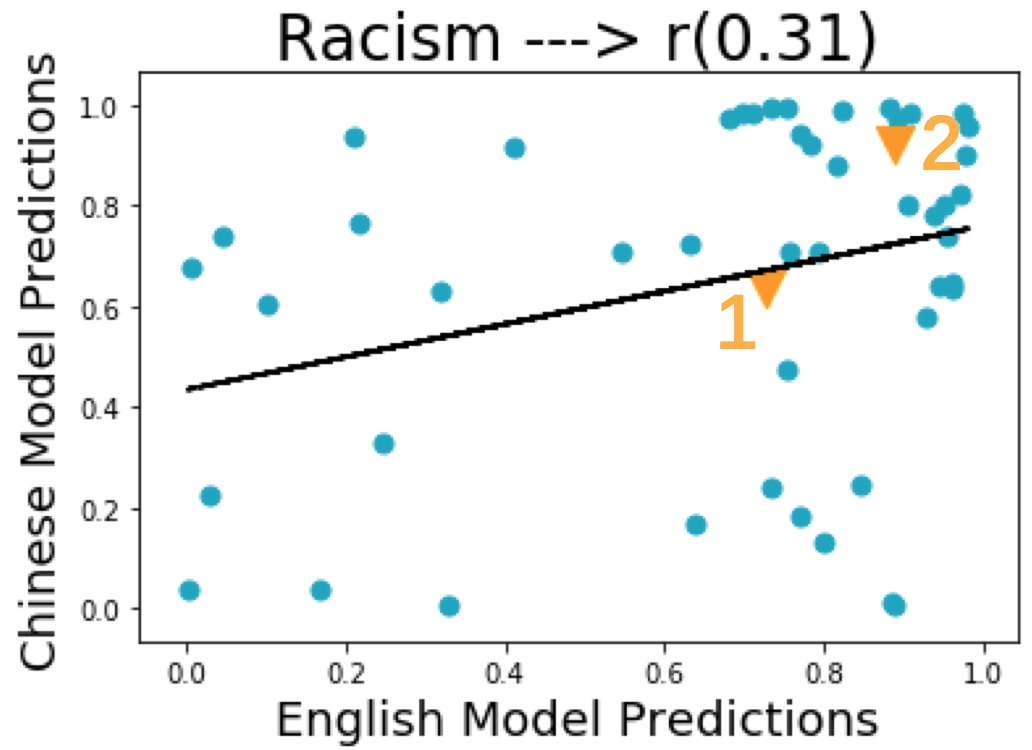}
\caption{Model predictions on \textit{racism} of English (x-axis) and Chinese (y-axis) models, and the correlation coefficient. The orange triangles represent cross-lingually agreed sentences: \textbf{1.} \textit{Racism is the prejudice against other cultures through identification of physical appearance and cues.} \textbf{2.} \textit{Fat shaming and/or body shaming can be just as bad as racism or homophobia.}}
\label{fig:savings}
\end{figure}

\section{Additional Details for Data Collection}
\subsection{Training Data}
We have described the procedure of collecting our training data from multi-lingual Wikipedia articles in Section \ref{sec:training-data}. In addition, for pre-processing details, we utilize Jieba\footnote{https://pypi.org/project/jieba/} and Mecab\footnote{https://pypi.org/project/mecab-python3/} to tokenize Chinese and Japanese sentences.

Back-translation (discussed in  Section \ref{sec:bias}) is the backbone of our CLUSTER model. Table \ref{table:BT_platforms} show the pivot languages as well as different translation systems used for our English, Chinese and Japanese models.

\begin{table}[H]
 \small
 \centering
\begin{tabular}{l|c|c}
\hline
         & Pivot language & Translation model       \\ \hline
English  & German         & \begin{tabular}[c]{@{}c@{}}Fined-tuned  transformer\\ \cite{ng2019facebook}
\end{tabular} \\ \hline
Chinese  & Japanese       & Youdao API              \\ \hline
Japanese & Chinese        & Google Translate        \\ \hline
\end{tabular}
\caption{The pivot languages and translation systems that we use for back-translation.}
\label{table:BT_platforms}
\end{table} 
\subsection{Questionnaire for Selecting Test Data}
\begin{figure*}[]
\centering
\includegraphics[width = 1.0\linewidth]{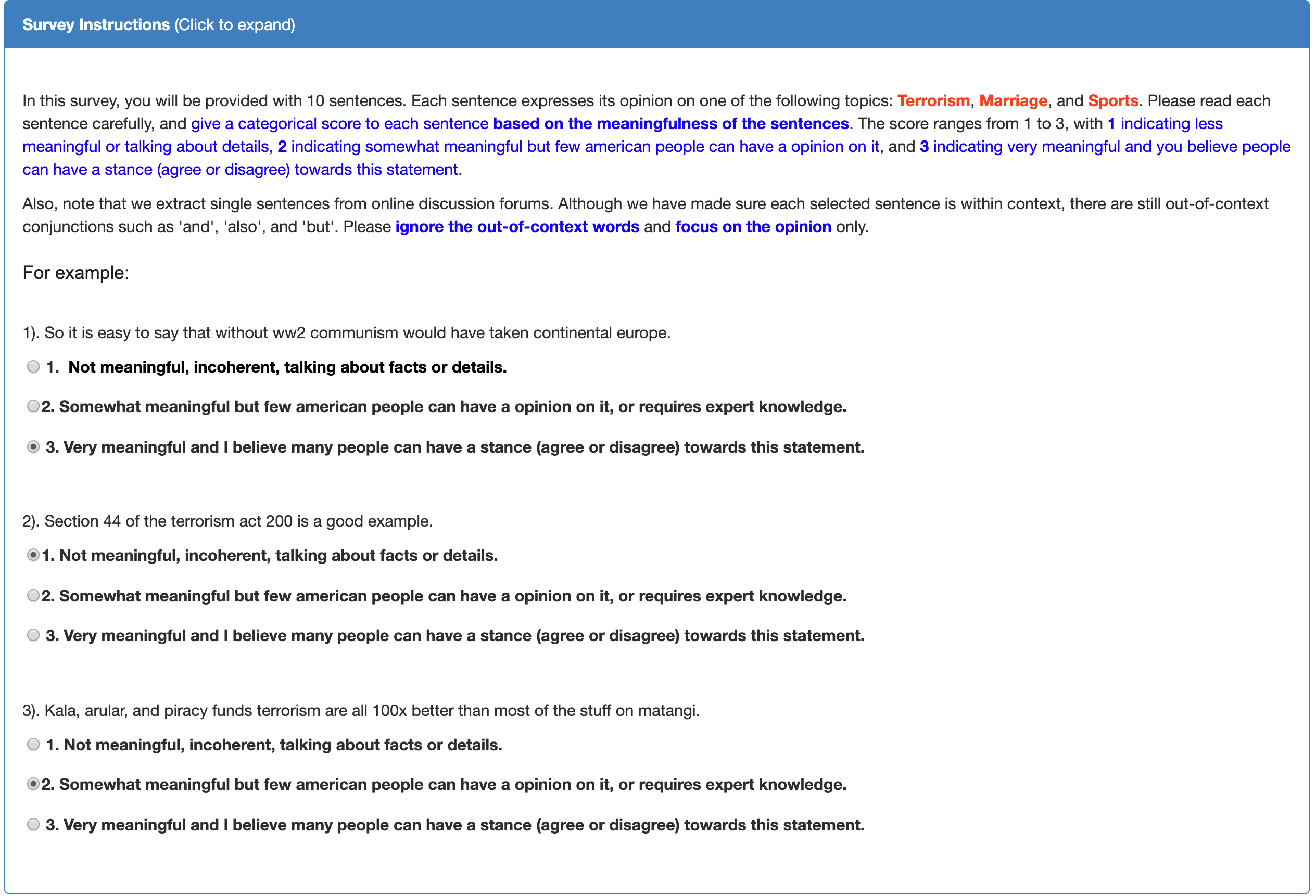}
\caption{The instructions of our survey to evaluate the meaningfulness of the IMO/IMHO sentences on the MTurk platform.}
\label{fig:survey-meaningful}
\end{figure*}

We design questionnaires to select out meaningful and high-quality claims from the original IMO/IMHO dataset (discussed in Section \ref{Section:TDS}), and collect three answers per claim.  Figure \ref{fig:survey-meaningful} shows our instructions to the English annotators on the Amazon Mechanical Turk (MTurk) platform. 

The turkers are asked to give a categorical score to each candidate sentence. The categorical score ranges from 1 to 3, with 1 indicating not meaningful, incoherent, or talking about facts, 2 indicating somewhat meaningful but few people have opinions on it, and 3 indicating highly meaningful. Because we extract single sentences from online discussion forums, we ask the turkers to ignore the out-of-context words such as `and', `also', and `but', and focus on the opinion only. Finally, if all annotators agree that a given claim is meaningful enough so that other people will hold a stance (either agreement or disagreement) towards it, we regard this candidate claim as one of our test samples for the final human annotation step.

\subsection{Questionnaire for Collecting Human Annotation}
\begin{figure*}[]
\centering
\includegraphics[width = 1.0\linewidth]{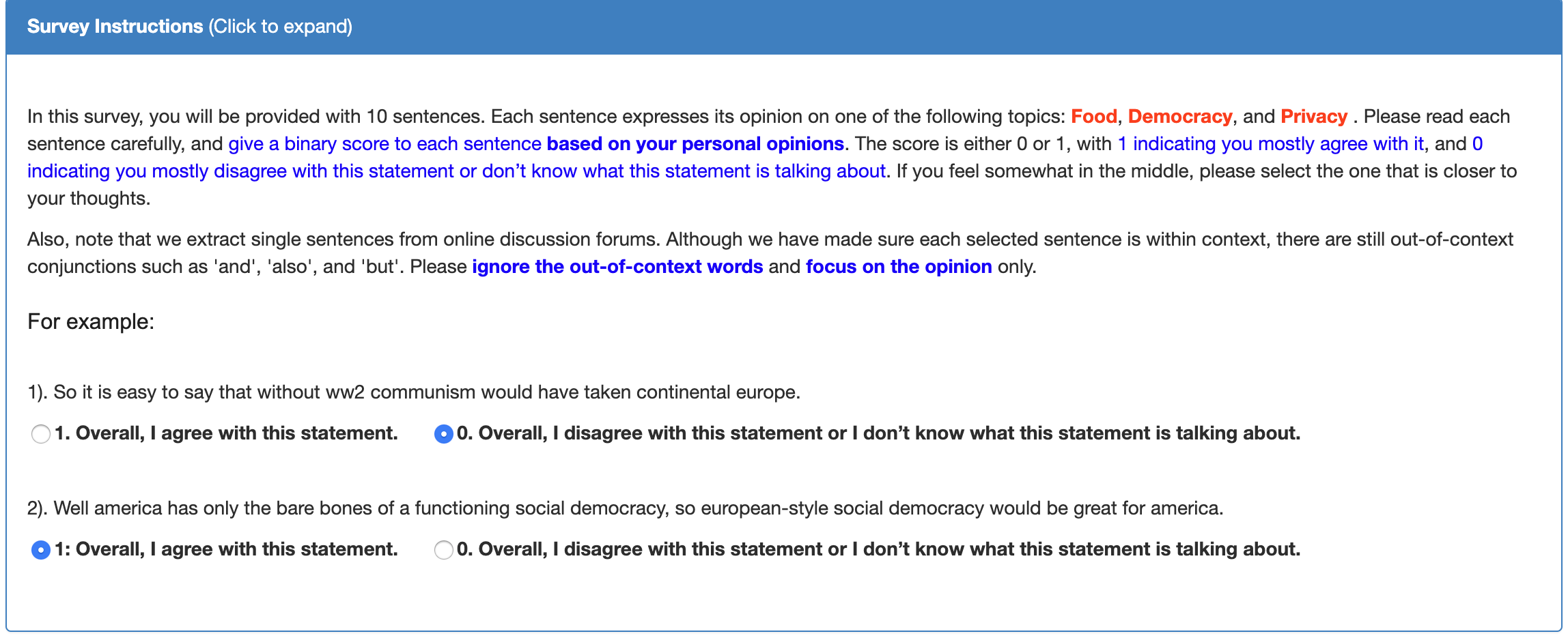}
\caption{An example instruction page of our survey to collect the human annotations on the MTurk platform.}
\label{fig:survey-opinion}
\end{figure*}

Figure \ref{fig:survey-opinion} is an English demonstration of our survey to collect human annotations of the test data. The annotators as instructed to read each sentence carefully, and give a binary score to each sentence based on their personal opinions. The score is either 0 or 1, with 1 indicating they mostly agree with this statement, and 0 indicating they mostly do not agree with it, or don’t know what this statement is talking about. 

Besides, we adopt attention checks to control the quality of our collected annotations. To this end, we manually select 7 facts from Wikipedia as attention check statements, which are obviously true to the masses, such as \textit{`Cheese is a dairy product derived from milk that is produced in a wide range of flavors, textures, and forms'}. We insert an attention check statement after every 9 test claims. If an annotator does not agree with one of our attention check statements, his entire HIT is rejected. Each annotator is allowed to annotate at most 20 sentences including the attention check statements.

\end{document}